\documentclass[runningheads]{llncs}
\usepackage[T1]{fontenc}
\usepackage{graphicx,verbatim}
\usepackage{booktabs}
\usepackage[colorlinks=true, linkcolor=blue, citecolor=green, urlcolor=red]{hyperref}

\usepackage{multirow}
\usepackage{amsmath}
\usepackage{bbding}
\usepackage{amsfonts}
\usepackage{subcaption} 

\begin{document}
\title{VBCD: A Voxel-Based  Framework for Personalized Dental Crown Design}

%


\author{Linda Wei\inst{1(\dagger)} 
\and Chang Liu\inst{2,5(\dagger)} 
\and Wenran Zhang\inst{4} 
\and Zengji Zhang\inst{6} 
\and Shaoting Zhang\inst{5} 
\and Hongsheng Li\inst{1,3(*)}
}

\authorrunning{Wei et al.}
     
\institute{${ }^1$ Multimedia Laboratory (MMLab), The Chinese University of Hong Kong\\ ${ }^2$ College of Biomedical Engineering, Fudan University \\
${ }^3$ Centre for Perceptual and Interactive Intelligence (CPII) under InnoHK
\\
${ }^4$ Department of Second Dental Center,  Shanghai Ninth People’s Hospital, Shanghai Jiao Tong University School of Medicine \\ 
${ }^5$ Sensetime Research \\
${ }^6$ School of Biomedical Engineering, Shanghai Jiao Tong University \email{1155230127@link.cuhk.edu.hk, 23110720100@m.fudan.edu.cn, hsli@ee.cuhk.edu.hk}}

\begingroup
\renewcommand\thefootnote{$\dagger$}
\footnotetext{Equal contribution.}
\endgroup
\begingroup
\renewcommand\thefootnote{*}
\footnotetext{Corresponding author.}
\endgroup
\begingroup
\renewcommand\thefootnote{**}
\footnotetext{This paper has been accepted in MICCAI 2025.}
\endgroup
\maketitle             

%
\begin{abstract}

The design of restorative dental crowns from intraoral scans is labor-intensive for dental technicians. To address this challenge, we propose a novel voxel-based framework for automated dental crown design (VBCD). The VBCD framework generates an initial coarse dental crown from voxelized intraoral scans, followed by a fine-grained refiner incorporating distance-aware supervision to improve accuracy and quality. During the training stage, we employ the Curvature and Margin line Penalty Loss (CMPL) to enhance the alignment of the generated crown with the margin line. Additionally, a positional prompt based on the FDI tooth numbering system is introduced to further improve the accuracy of the generated dental crowns. Evaluation on a large-scale dataset of intraoral scans demonstrated that our approach outperforms existing methods, providing a robust solution for personalized dental crown design. 
The related code is in: \url{https://github.com/lullcant/VBCD}


\end{abstract}
\section{Introduction}

\begin{figure}[t]
    \centering
\includegraphics[width=\linewidth]{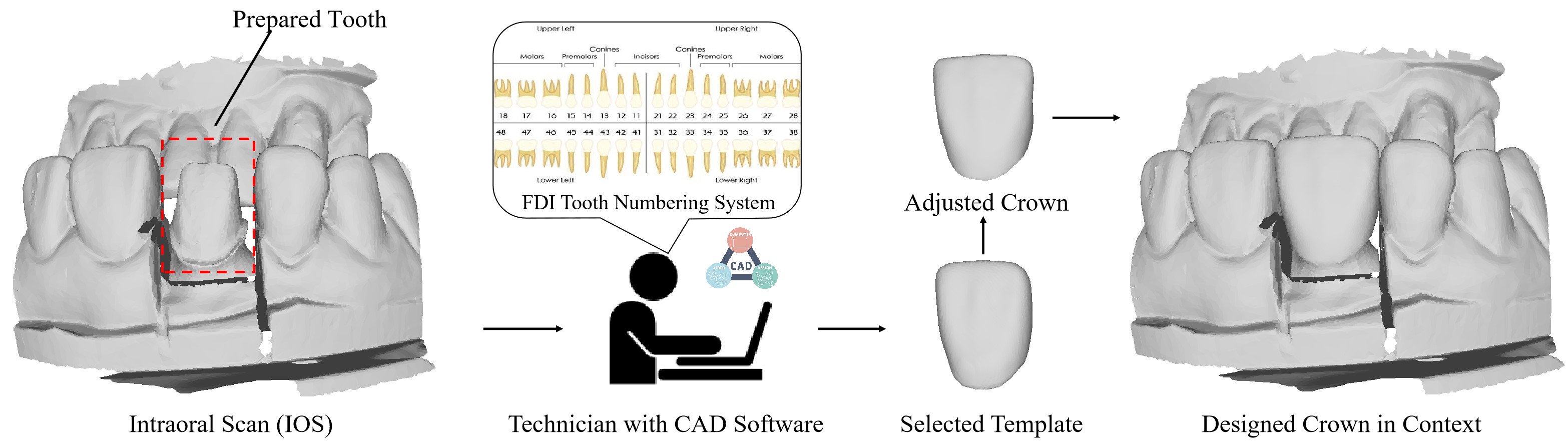}
    \caption{The CAD workflow for dental crown design initiates with the acquisition of an intraoral scan, upon which the technician adjusts the template crown selected according to Fédération Dentaire Internationale (FDI) tooth numbering system. }
    \label{fig:intro}
\end{figure}

Oral diseases are among the most common non-communicable diseases worldwide, affecting an estimated
3.5 billion people \cite{world2022global}. Tooth injuries due to wear, caries, or trauma are the most prevalent forms of dental malfunction. For patients with abraded teeth, prosthodontics, specifically crown restoration, is a typical preferred treatment option \cite{raigrodski2012survival}. Therefore, designing an artificial crown to restore dental function is critical to clinical dental practice. Modern dental computer-aided design (CAD) systems have greatly advanced the design of dental crowns \cite{davidowitz2011use,jain2016cad,miyazaki2009review}. As illustrated in \autoref{fig:intro}, dental technicians select an appropriate crown template for the prepared tooth according to the Fédération Dentaire Internationale (FDI) tooth numbering system \cite{fdi2009dentistry}, which is then modified based on the patient’s intraoral scan (IOS) to fabricate a restoration crown.

Despite their advantages, the workflow of CAD systems for crown restoration remains labor-intensive due to the lack of customization of crown templates. Dental technicians are required to make extensive and intricate modifications to the templates of target teeth, carefully adjusting them to accommodate the positions and occlusal relationships of adjacent and opposing teeth \cite{turkyilmaz2021tooth}. Therefore, developing an automated method capable of generating customized crown templates or final crowns is essential to alleviate the workload of dental technicians.

Recently, with the development of artificial intelligence for clinical application \cite{wang2024sensecare}, the use of deep learning techniques for prosthesis design has emerged as a promising avenue for exploration. For instance, 3D-CNN and GANs have been employed to generate partial dental crowns and occlusal surface reconstructions \cite{farook2023computer,feng20233d,tian2022dual}. However, these approaches often suffer from low accuracy and a lack of customization, primarily due to their inability to account for antagonist teeth.
Recent studies \cite{hosseinimanesh2024personalized,hosseinimanesh2023mesh,yang2024dcrownformer} have framed the task of customized dental crown generation as a mesh completion problem, whose objective is to generate a dental crown mesh that restores the abraded areas with the information from the IOS. Most of the deep neural network models are not well-suited for handling mesh data, so the mesh completion task is usually decomposed into two main steps: point cloud completion and mesh reconstruction.

Although many studies have demonstrated excellent performance in point cloud completion \cite{tchapmi2019topnet,xie2020grnet,yuan2018pcn}, most of them do not take normal vectors of the completed points into account, which are essential for many mesh reconstruction algorithms. 
To solve this limitation, Shape as Points (SAP) \cite{peng2021shape} incorporates a novel module that combines an encoder-decoder model for normal vector prediction with a Differentiable Poisson Surface Reconstruction (DPSR) module. This module facilitates the reconstruction of the mesh surface from the point cloud by constructing a Poisson Surface Reconstruction (PSR) indicator function grid and applying the Marching Cubes algorithm \cite{lorensen1998marching} to it.

Inspired by this study, several works \cite{hosseinimanesh2024personalized,hosseinimanesh2023mesh,yang2024dcrownformer} achieved end-to-end mesh completion for the customized dental crown by combining a point cloud completion network with SAP.
Despite their promising performance, there are two primary limitations in previous methods. \textbf{First}, 
the number of points in the generated crown mesh was fixed and limited, while it varies with the type of tooth in clinical practice.
\textbf{Second}, most studies were conducted on small-scale datasets or only focused on molars. The evaluation of the robustness and generalization was insufficient.

In this paper, we propose \textbf{V}oxel-\textbf{B}ased network for dental \textbf{C}rown \textbf{D}esign (VBCD) to address these issues. We first convert the input IOS into volume and utilize a 3D UNet backbone to generate a coarse dental crown in the volume modality. To mitigate the loss of geometric details during the voxelization process, we adopt a coarse-to-fine strategy to refine the coarse dental crown, incorporating a distance-based loss function to further improve the performance. Additionally, we introduce a positional prompt based on the FDI tooth numbering system to improve the customization of the generated crown. Our contributions are summarized as follows:

\begin{itemize} 
\item 

We propose VBCD, a coarse-to-fine framework that automatically generates personalized crowns based on an intraoral scan with a prepared tooth. The framework can process an arbitrary number of input and output points, which facilitates the generation of accurate and fine-grained dental crowns. 

\item We design a tooth position prompt based on the FDI tooth numbering system and utilize the Curvature and Margin Penalty Loss (CMPL) to further enhance the performance of the crown generation framework. 

\item We construct the most extensive oral scan dataset on dental crown generation tasks to date, encompassing a complete set of tooth types, and perform extensive experiments to demonstrate the robustness and generalization of our framework utilizing the dataset.

\end{itemize}
\section{Methods}
\begin{figure}[t]
    \centering
    \includegraphics[width=\textwidth]{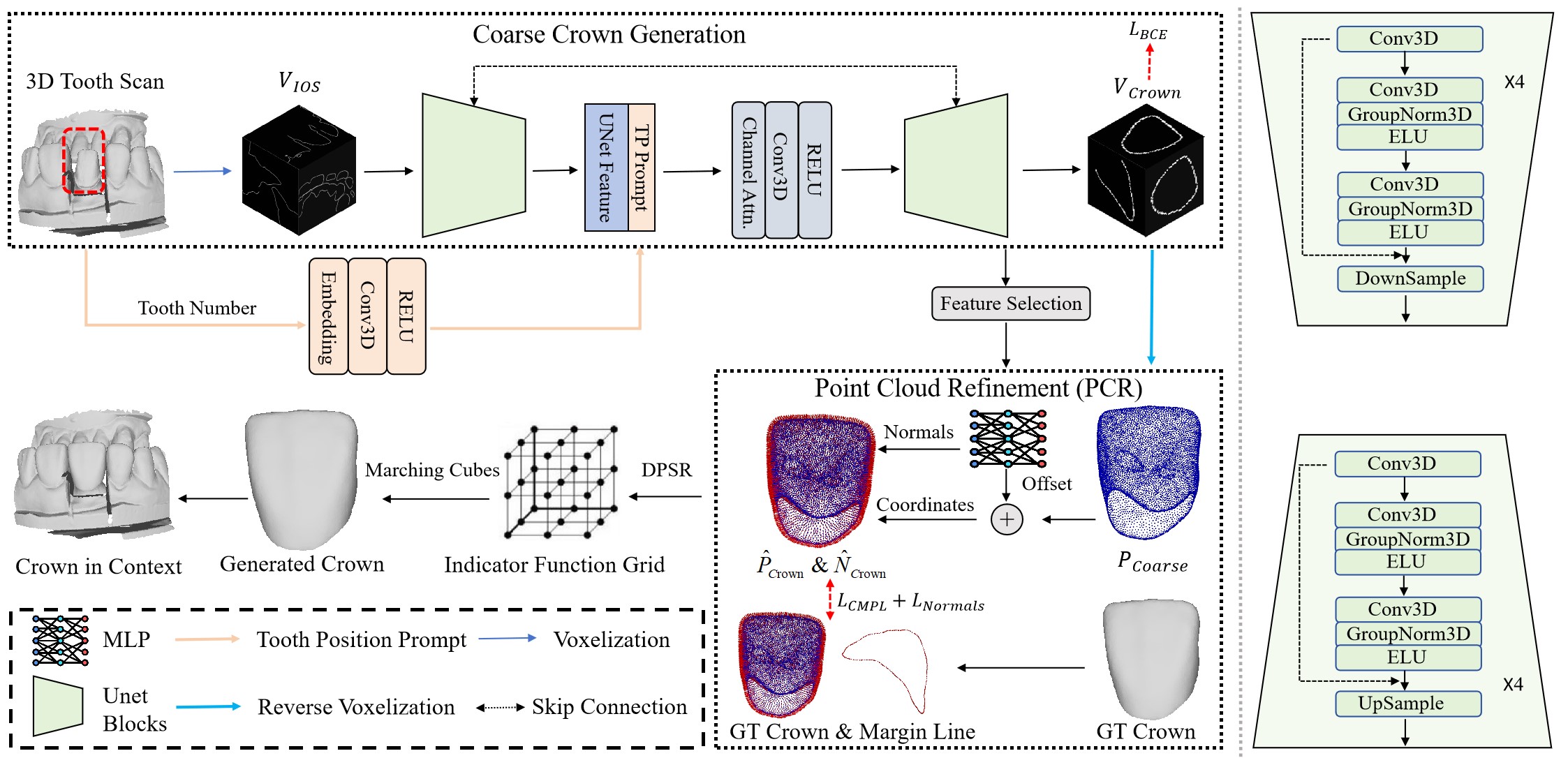}
    \caption{Overall architecture of our VBCD. Given an IOS with a prepared tooth, our framework first generates a coarse crown in volume modality, then further refines the coarse result in point cloud representation through PCR.}
    \label{fig:architecture}
\end{figure}

The overall architecture of VBCD is shown in \autoref{fig:architecture}.  The framework follows a coarse-to-fine paradigm to generate fine-grained dental crowns according to the IOS. The coarse dental crown is initially generated in volume modality. Further optimization is performed on this structure with a point cloud representation.

\subsection{Voxelization}
The input of our framework is an IOS with the prepared tooth to be restored. To regularize the point cloud input and make our framework adaptable to an arbitrary number of input points, we transform the IOS point cloud data into a volume, which is a structured representation with spatial context for CNN. The given IOS is voxelized as a volume $V_{\text{IOS}} \in \mathbb{R}^{D\times H \times W}$  with the spacing $s$. The origin of the IOS bounding box $(o_x,o_y,o_z)$ is also set to be the volume origin. For each point $(x,y,z)$ in the IOS, the value of the corresponding voxel $V_{\text{IOS}}^{(i,j,k)}$ is set to be 1, where the index of voxel is $(i,j,k)= \lfloor \frac{(x,y,z)-(o_x,o_y,o_z)}{s} \rfloor$.

For volume ground truth, the dental crown is voxelized as a volume $V_{\text{GT}}$ by applying the same procedure as $V_{\text{IOS}}$. 

\subsection{Coarse Crown Generation}
Volumes voxelized from the IOS data are fed into a 3D UNet backbone for a coarse crown generation. We introduce a tooth position (TP) prompt to incorporate the context of tooth position. The label of the prepared tooth in the FDI tooth numbering system \autoref{fig:intro} is explicitly encoded as an embedding of $128$ dims and then concatenated to the output of the bottleneck feature of UNet. Subsequently, a channel attention module \cite{wang2020eca} aggregates the TP prompt and the UNet feature, and a convolution layer is applied to ensure the compatibility of the feature with the shape required by the decoder.

The final output $f \in \mathbb{R}^{C\times D\times H \times W}$ of the backbone is fed into a convolution layer $Conv_{\text{coarse}}$. The output of $Conv_{coarse}$, which is the logits of crown volume, is denoted as  $\hat{\mathbf{L}} \in \mathbb{R}^{1\times D \times H \times W}$. We apply BCE loss to supervise the coarse crown generation, where $\sigma$ is the sigmoid function.
$$
\mathcal{L}_{BCE} = - \left( V_{GT} \cdot \log(\sigma(\hat{\mathbf{L}})) + (1 - V_{GT}) \cdot \log(1 - \sigma(\hat{\mathbf{L}})) \right)
$$

\noindent The predicted coarse crown volume $V_{\text{Crown}} \in \{0,1\}^{D \times H \times W}$ is a binary volume derived by thresholding $\hat{\mathbf{L}}$, setting voxels greater than zero to one. We obtain the points on the coarse crown, denoted as $P_{\text{coarse}} \in \mathbb{R}^{N\times 3} $, through reverse voxelization of $V_{\text{Crown}}$.

\subsection{ Crown Generation Refinement}
Although we can get a coarse generation result with the procedure above, there are two major issues that remain in the coarse crown generation: 
\textbf{1.} Voxelization inevitably results in the loss of fine geometric details from the original mesh point.
\textbf{2.} BCE loss is a voxel-level loss that lacks the ability to provide distance-aware supervision.
To address these issues, we introduced a point cloud refiner (PCR) to further refine the generated coarse crown $P_{\text{coarse}}$. 
The feature embedding $e \in \mathbb{R}^{N\times C}$ of the coarse crown point is gathered from the final layer feature $f$ of UNet by the mask selection procedure:

$$
 \mathbf{F} = f\odot \mathbf{M}\in \mathbb{R}^{C\times D\times H \times W} , \quad
 e = \text{Flatten}(\mathbf{F}[\mathbb{I}_{\{\mathbf{F}\neq 0\}}])\in \mathbb{R}^{N\times C}, 
$$
in which $\mathbf{M} \in \{0,1\}^{C\times D\times H \times W}$ is the mask obtained from broadcasting  $V_{\text{Crown}}$ in the channel dimension. $N$ is the number of points on the coarse crown, and $\odot$ is the Hadamard Product. $e$ is employed to predict the offset between $P_{\text{Coarse}}$ and the points in the ground truth dental crown $P_{GT}$. The feature embedding is also utilized to predict the normal vector of each point, which is indispensable for reconstructing the crown mesh. 

The fine-grained dental crown point cloud $\hat{P}_{\text{crown}}$ and the normal vectors $\hat{N}_{\text{crown}}$ are computed using the following formula:
$$\hat{P}_{\text{crown}} = P_{\text{coarse}}+\text{MLP}_1(e),\quad \hat{N}_{\text{crown}} = \text{MLP}_2(e)$$

\noindent Motivated by \cite{yang2024dcrownformer}, we introduce the Curvature and Margin Penalty Loss (CMPL), which better assists the model in generating crown details and ensuring more accurate delineation of the margin line, to supervise the fine-grained prediction. The CMPL is formulated as follows:

\begin{equation*}
\begin{aligned}
\mathcal{L}_{\text{CMPL}} = \frac{1}{|\hat{P}_{\text{Crown}}|}\sum_{\mathbf{p} \in \hat{P}_{\text{Crown}}} \left( e^{|\kappa(\mathbf{p})|} + \mathbb{I}_{\{\mathbf{p} \in M(P_{\text{GT}})\}} \right) \min_{\mathbf{q} \in P_{\text{GT}}} \|\mathbf{p} - \mathbf{q}\|_2  \\
+ \frac{1}{|P_{\text{GT}}|} \sum_{\mathbf{q} \in P_{\text{GT}}} \left( e^{|\kappa(\mathbf{q})|} + \mathbb{I}_{\{\mathbf{q} \in M(P_{\text{GT}})\}} \right) \min_{\mathbf{p} \in \hat{P}_{\text{Crown}}} \|\mathbf{p} - \mathbf{q}\|_2
\end{aligned}
\end{equation*}
where $\kappa(\mathbf{p})$ is the curvature of point $\mathbf{p}$; $M(P_{GT})$ is the margin line points of $P_{GT}$, which is illustrated in \autoref{fig:architecture}.
The normal vector $\hat{N}_{\text{crown}}$ of each point is supervised by the MSE loss denoted as $\mathcal{L}_{\text{Normals}}$. The ground truth normal vector of each point in the generated crown is defined as the normal vector of its nearest neighbor in the ground truth.

\noindent The overall loss of the proposed method is formulated as:
$$
\mathcal{L}_{\text{total}} = \mathcal{L}_{\text{BCE}} +\mathcal{L}_{\text{CMPL}} + \mathcal{L}_{\text{Normals}}
$$
With the generated points and the corresponding normal vectors, it's handy to reconstruct the mesh of the dental crown with some plug-and-play surface reconstruction algorithms \cite{huang2023neural,kazhdan2006poisson,kazhdan2013screened}. In this paper, we use the DPSR and Marching Cubes algorithms, which were also utilized by previous research, for fair comparison.
Specifically, we first utilize the DPSR module in SAP to estimate an indicator function grid by solving the Poisson equation with the point cloud and its normals. The iso-surface from this indicator function grid is extracted as the reconstructed mesh with the marching cubes algorithm \cite{lorensen1998marching}.
\section{Experiments}
\begin{figure}[!h]
    \centering
    \includegraphics[width=\linewidth]{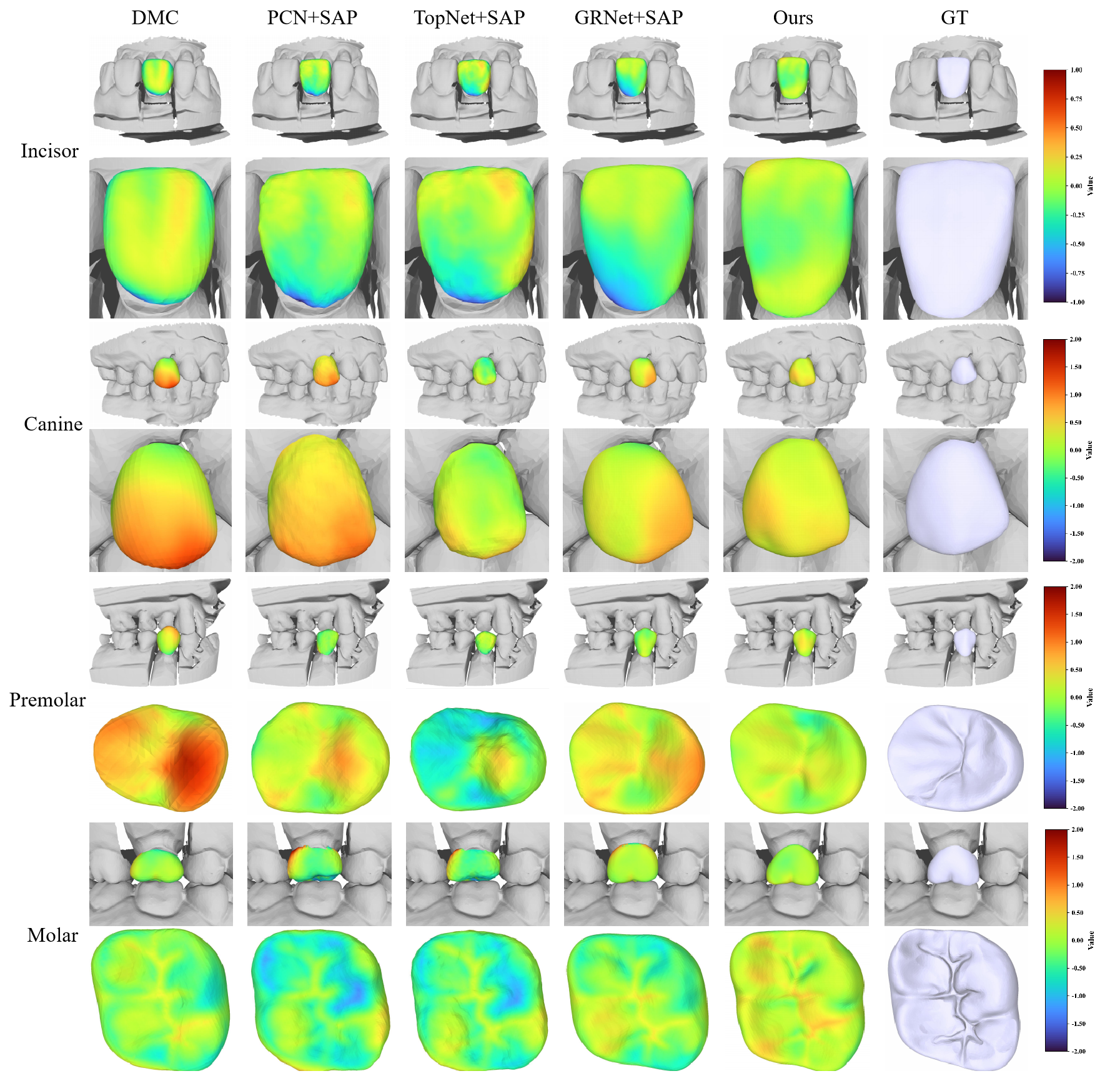}
    \caption{Comparison Experiment Results. The color map denotes the distance between a point in the generated crown and its nearest neighbor in the ground truth. No significant compromise of the occlusal relationship was detected.}
    \label{Comparison}
\end{figure}

\subsection{Dataset and Implementation Details}
\noindent \textbf{Dataset.} 
Our dataset comprises 6,499 oral scans with single-tooth edentulous, covering all the tooth types, including incisors, canines, premolars, and molars. The corresponding restoration crown of each oral scan is also included. For each oral scan, a 2cm × 2cm × 2cm cubic region centered on the crown's center is extracted. The margin line of the crown is defined as all the edges that belong to only one face in the crown mesh. The dataset is split into a training, validation, and test set with a ratio of 7:1:1. We employ stratified sampling to handle different tooth types, ensuring that the distribution of tooth types in the training and test sets remains consistent.

\noindent \textbf{Implementation Details.}
Our framework was implemented using PyTorch and trained on 2 NVIDIA GTX 4090 GPUs with a batch size of 4 for 720000 iterations. AdamW was used as the optimizer, and the initial learning rate was set to be $10^{-4}$. The inference time is about 357ms/case, which is much more efficient compared to the manual design (5-10 min/case)\cite{miyazaki2009review}. To minimize the quantization effects of the voxelization and keep geometric details \cite{xie2020grnet}, the input intraoral scan was voxelized as a \( 128^3 \) volume with a fine-grained spacing \( 0.15 \text{mm} \). The UNet backbone contained 4 downsampling and upsampling blocks; the base channel of feature $C$ was 64. To enhance the stability of the training process, we employed a two-stage training strategy. During the initial stage, only \(L_{\text{coarse}}\) was applied as supervision. The $\mathcal{L}_{\text{CMPL}}$ and $\mathcal{L}_{\text{Normals}}$, which were designed for crown refinement, were incorporated in optimization after 400,000 iterations.

\begin{table}[!t]
\centering
\caption{Experimental Results: (a) Comparison Experiment Results on All the Tooth Types, (b) Ablation Study on Tooth Numbering Prompt and CMPL}
\label{tab:combined_results}
\begin{subtable}[t]{\textwidth}
\centering
\caption{Comparison Experiment Results on All the Tooth Types.}
\label{tab:comparison}
\begin{tabular}{ccccccc}
\toprule
\textbf{Metric} & \textbf{Method} & \textbf{Incisor} & \textbf{Canine} & \textbf{Premolar} & \textbf{Molar} & \textbf{Overall} \\
\midrule
\multirow{5}{*}{CD-L2 $\downarrow$ ($mm^2$)} 
    & DMC \cite{hosseinimanesh2023mesh} & 0.390 &0.621 & 0.363 & 0.362 & 0.375 \\
    & PCN+SAP \cite{yuan2018pcn} & 0.367 & 0.471 & 0.345 & 0.347 & 0.354 \\
    & TopNet+SAP \cite{tchapmi2019topnet} & 0.505 & 0.576 & 0.503 & 0.532 & 0.523 \\
    & GRNet+SAP  \cite{xie2020grnet} & 0.300 & 0.328 & 0.288 & 0.285 & 0.290 \\
    & Ours & \textbf{0.161} & \textbf{0.177} & \textbf{0.138} & \textbf{0.133} & \textbf{0.140} \\
\midrule
\multirow{5}{*}{Fidelity  $\downarrow$ ($mm$)}
    & DMC \cite{hosseinimanesh2023mesh} & 0.361 & 0.458 & 0.363 & 0.384 & 0.377 \\
    & PCN+SAP \cite{yuan2018pcn} & 0.335 & 0.358 & 0.318 & 0.336 & 0.332 \\
    & TopNet+SAP \cite{tchapmi2019topnet} & 0.386 & 0.403 & 0.396 & 0.416 & 0.405 \\
    & GRNet+SAP  \cite{xie2020grnet} & 0.273 & 0.267 & 0.258 & 0.280 & 0.273 \\
    & Ours & \textbf{0.217} & \textbf{0.225} & \textbf{0.216} & \textbf{0.210} & \textbf{0.213} \\
\midrule
\multirow{5}{*}{F-score $\uparrow$}
    & DMC \cite{hosseinimanesh2023mesh} & 0.760 & 0.631 & 0.747 & 0.818 & 0.785 \\
    & PCN+SAP \cite{yuan2018pcn} & 0.817 & 0.759 & 0.816 & 0.870 & 0.845 \\
    & TopNet+SAP \cite{tchapmi2019topnet} & 0.744 & 0.708 & 0.700 & 0.769 & 0.745 \\
    & GRNet+SAP \cite{xie2020grnet} & 0.905 & 0.866 & 0.901 & 0.934 & 0.918 \\
    & Ours & \textbf{0.928} & \textbf{0.928} & \textbf{0.949} & \textbf{0.970} & \textbf{0.957} \\
\bottomrule
\end{tabular}
\end{subtable}
\begin{subtable}[t]{\textwidth}
\centering
\caption{Ablation Study (only overall metrics are provided because of limited page count).}
\label{ablation}

\begin{tabular}{cccccc}
\toprule
\multicolumn{3}{c}{\textbf{Components}} & \multirow{2}{*}{\textbf{CD-L2 (mm) $\downarrow$}} &  \multirow{2}{*}{\textbf{Fidelity (mm) $\downarrow$}} &  \multirow{2}{*}{\textbf{F-Score $\uparrow$}} \\

\textbf{PCR} & \textbf{TP Prompt} & \textbf{CMPL} & & & \\
\midrule
 \XSolidBrush  &\XSolidBrush & \XSolidBrush & 0.230 & 0.314 & 0.896 \\
\Checkmark & \XSolidBrush  & \XSolidBrush  & 0.198 & 0.231 & 0.929 \\
\Checkmark & \XSolidBrush & \Checkmark & 0.154 & 0.216 & 0.934 \\
\Checkmark & \Checkmark & \XSolidBrush  & 0.156 & 0.219 & 0.932 \\
\Checkmark & \Checkmark & \Checkmark & \textbf{0.140} & \textbf{0.213} & \textbf{0.957} \\
\bottomrule
\end{tabular}
\end{subtable}

\end{table}

\subsection{Results}
\subsubsection{Evaluation Metrics.}
\noindent To better assess the performance, all the distance-related metrics are calculated in the \textbf{physical coordinate}, which is measured in \textbf{millimeters (mm)}. 
We use L2 chamfer distance (CD-L2) and F-score to measure the similarity between the generated and the ground truth point cloud, as in the previous works \cite{hosseinimanesh2023mesh,yang2024dcrownformer}. Moreover, to better evaluate the consistency of the generated crown to the ground and reduce the impact of artifacts, we adopted fidelity \cite{yuan2018pcn} as an additional evaluation metric.

\noindent \textbf{Comparison Experiment.}
We compared our method with the open-sourced crown generation method DMC \cite{hosseinimanesh2023mesh} and some widely used point cloud completion networks with SAP. Notably, SAP is required for all methods as it generates the PSR indicator function grid necessary for mesh reconstruction. The qualitative results of the comparative experiments are presented in \autoref{tab:comparison}. Our method outperforms previous approaches across all tooth types and evaluation metrics. For tooth types with limited data, such as canines, the performance degradation is minimal, demonstrating the robustness of our approach. The visualization results, shown in \autoref{Comparison}, indicate that our model exhibits higher similarity to the ground truth and produces more finely detailed crowns with greater precision than previous methods (\autoref{Comparison} rows 2, 4, 6, 8). Specifically, the margin lines of the crowns generated by our method are significantly more precise, which is critical for clinical applications (\autoref{Comparison} rows 2, 4, 5, 7).

\noindent \textbf{Ablation Study.}
We performed ablation studies in \autoref{ablation} to verify the effectiveness of our design. The baseline uses the UNet backbone to generate a restoration crown (point coordinates and normal vectors) directly. We evaluated the effect of components in our design, including (1) PCR, (2) TP Prompt, and (3) CMPL. For the model without CMPL, we used CPL in \cite{yang2024dcrownformer} as a distance-aware loss. In \autoref{ablation}, there is a performance improvement with each component added to the baseline. The VBCD, which is the ensemble of all the components and baseline, outperformed all the other experiment settings.

\label{tab:comparision}

\section{Conclusion}
In this paper, we propose a coarse-to-fine framework for generating personalized dental crowns across all tooth types. Our framework leverages voxelization to regularize unordered point clouds and further incorporates CMPL and tooth position prompts to enhance the precision, detail, and fit of the margin lines of the generated crowns. Quantitative experiments and visualization results demonstrate that our method outperforms previous approaches across all metrics and is adaptable to all tooth types. However, our framework is limited by high memory consumption due to high-resolution voxelization \cite{fei2022comprehensive}. Future work could address this issue by incorporating sparse convolutions \cite{choy20194d,liu2015sparse} as the encoder in the UNet, improving computational efficiency.
\subsection*{Disclosure of Interest}
The authors have no competing interests to declare that
are relevant to the content of this article.

\bibliographystyle{splncs04}
\bibliography{parts/ref}

\end{document}